\pdfoutput=1
\documentclass{article}

\usepackage[preprint]{neurips_2026}

\usepackage[utf8]{inputenc} 
\usepackage[T1]{fontenc}    
\usepackage{hyperref}       
\usepackage{url}            
\usepackage{booktabs}       
\usepackage{amsmath}       
\usepackage{amsfonts}       
\usepackage{nicefrac}       
\usepackage{microtype}      
\usepackage{xcolor}         
\usepackage{multirow}
\usepackage{colortbl}
\usepackage{graphicx}
\usepackage{wrapfig}
\usepackage{xspace}
\usepackage{algorithm}
\usepackage{algpseudocode}
\hypersetup{hidelinks,hypertexnames=false,pdftitle={Verifiable Geometry Problem Solving: Solver-Driven Autoformalization and Theorem Proposing},pdfauthor={lican and zhangting}}
\newcommand{\ours}{\textsc{SD-GPS}\xspace}
\newcommand{\backbone}{QwenVL3-2B\xspace}

\title{Verifiable Geometry Problem Solving: Solver-Driven Autoformalization and Theorem Proposing}

\author{%
  Can Li 
  \quad
  Ting Zhang
  \quad
  Junbo Zhao
  \quad
  Hua Huang \\
  \textbf{Beijing Normal University}
}

\begin{document}

\maketitle
\begin{abstract}
Geometry Problem Solving have increasingly adopt the neuro-symbolic paradigm, combining neural intuition with symbolic rigor. However, current frameworks suffer from severe bottlenecks in two core stages: autoformalization, which treats multimodal translation as a static task decoupled from downstream solver compatibility, and theorem prediction, where solvers frequently hit a deductive impasse due to fixed rule libraries. To address these, we propose SD-GPS, a solver-driven framework that treats the symbolic solver as an execution oracle throughout both formalization and deduction. First, Solver-Driven Autoformalization unifies supervised formal-language adaptation and solvability-guided reinforcement learning into a single module built on QwenVL3-2B, making executability the central training signal. Second, Verified Theorem Proposing introduces an impasse-aware agent that proposes local auxiliary lemmas from current proof states, ensuring soundness by filtering all proposals through symbolic verification. Empirical evaluations on Geometry3K and PGPS9K demonstrate that SD-GPS consistently outperforms existing MLLM, neural, and neuro-symbolic methods across standard completion, multiple-choice, and cross-modal reference regimes, proving that closing the loop between multimodal perception and symbolic execution significantly improves geometric reasoning, offering profound insights into how neural agents can be grounded by formal systems to achieve verifiable problem-solving capabilities.

\end{abstract}
\section{Introduction}
\label{sec:intro} 

Geometry Problem Solving (GPS) aims to derive mathematical solutions from a textual problem description and its corresponding diagram~\citep{chen2022geoqageometricquestionanswering,10.1145/1460361.1460381, sachan-xing-2017-learning, 10.1007/BF02328447,wu2024egps,peng2023geodrl}.
To tackle the inherent complexity of geometry, many recent systems adopt the neuro-symbolic paradigm~\citep{alphageometry,chervonyi2025goldmedalistperformancesolvingolympiad}. This dual-process approach utilizes neural networks as a "perceptual front-end" and symbolic engines as a "logical back-end"~\citep{wu2022autoformalizationlargelanguagemodels}. Within this framework, the reasoning pipeline is driven by two core modules: (i) Autoformalization, which translates multimodal raw inputs into executable formal representations, and (ii) Theorem Prediction, which selects or proposes theorem instances that can be used by the solver to derive solutions. 
This modular architecture allows systems to leverage the intuitive recognition of deep learning alongside the deterministic rigor of formal logic.

In the autoformalization stage, the objective is to transform multimodal inputs into solver-ready predicates. However, existing GPS frameworks typically adopt a “decoupled-and-rectified” paradigm~\citep{lu2021intergps, Zhao_2025_ICCV, autogps}, in which isolated neural modules independently parse diagrams and formalize text, followed by a post-hoc rectification step to resolve referential ambiguities~\citep{Zhao_2025_ICCV, autogps}. Beyond this structural fragility, a more fundamental limitation lies in objective misalignment. Current approaches treat formalization as a static, one-way semantic translation task, optimizing against expensive human-annotated logical forms that emphasize linguistic fidelity rather than computational solvability. 
As a result, the formalization process is not aligned with the ultimate goal of solver-based problem solving, and high semantic accuracy does not necessarily translate into effective or executable representations for downstream reasoning.

Even when provided with accurate formalization, the theorem prediction stage faces a second critical bottleneck: a fixed theorem library and finite search budget can leave the solver at a deductive impasse. State-of-the-art GPS frameworks~\citep{autogps,zhang2024formalgeoextensibleformalizedframework} usually operate with a pre-defined library of geometric rules, so they may fail when a problem requires a rarely selected theorem instance, a small auxiliary construction, or an intermediate relation such as cyclicity, similarity, or proportionality. We therefore treat theorem proposing as a bounded, verifier-controlled proposal problem rather than as free-form axiom generation: a neural agent may suggest local auxiliary lemmas from the current proof state, but only the symbolic back-end can accept, reject, and use them.

\begin{wrapfigure}{r}{0.36\textwidth}
    \vspace{-10pt}
    \centering
    \includegraphics[width=0.35\textwidth]{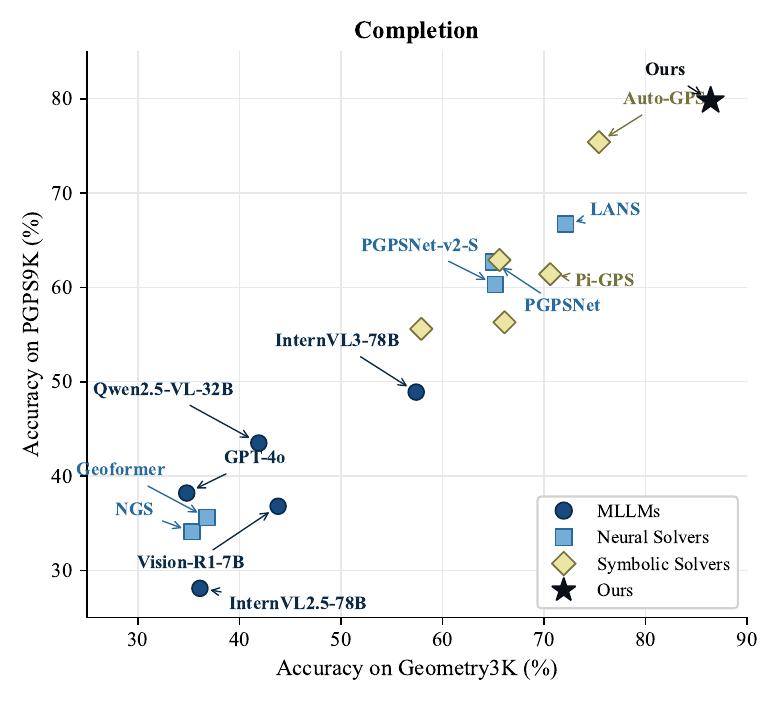}
    \vspace{-8pt}
    \caption{Performance comparison among existing methods.}
    \label{fig:performance_comparison}
    \vspace{-10pt}
\end{wrapfigure}

To resolve these misalignments, we propose \textsc{SD-GPS}, a solver-driven framework that treats the symbolic solver as an execution oracle throughout both formalization and deduction. First, Solver-Driven Autoformalization replaces the decoupled pipeline with a unified multimodal formalizer built on QwenVL3-2B. It reads the raw diagram and problem text jointly, learns the target formal language through supervised adaptation, and is further optimized with solvability-guided feedback from the solver. This makes executability, rather than surface-level textual similarity alone, the central training signal. Second, Solver-Driven Theorem Proposing addresses solver-side impasses. Instead of allowing a neural model to add unchecked axioms, the agent proposes bounded auxiliary lemmas or theorem instantiations from the current proof state, and every accepted proposal must pass symbolic verification before it can drive the derivation.


Empirical evaluations on Geometry3K~\cite{lu2021intergps} and PGPS9K~\cite{zhang2023pgpsnet} show consistent improvements over MLLM, neural, and neuro-symbolic baselines. As summarized in Figure~\ref{fig:performance_comparison}, \ours{} improves completion accuracy on both benchmarks over the strongest listed prior systems. The results suggest that closing the loop between multimodal perception and symbolic execution improves both solver-ready formalization and final geometric reasoning.

In summary, our contributions are three-fold:

\begin{itemize}

    \item 
\textbf{Solver-Driven Autoformalization:} We unify supervised formal-language adaptation and solvability-guided reinforcement learning into a single autoformalization module that produces solver-executable predicates from raw diagram-text inputs.
    \item 
\textbf{Verified Theorem Proposing:} We introduce an impasse-aware theorem-proposing agent that proposes auxiliary lemmas from solver states, while preserving soundness by accepting only proposals verified by the symbolic back-end.
    \item 
\textbf{Systematic Evaluation:} We evaluate under standard completion and choice settings, cross-modal point-reference regimes, and OCR-induced failure modes, showing robust gains on Geometry3K and PGPS9K.

\end{itemize}

\noindent
\section{Related Work}
\label{sec:related_work}

\noindent 
\textbf{Geometry problem solving.}
The field of GPS has evolved from classical rule-based engines to contemporary neural-symbolic hybrids. Early symbolic methods, such as Wu’s Method~\citep{wu1978method} and Groebner Basis~\citep{10.1145/1088216.1088219}, established a foundation of mathematical rigor but were fundamentally limited by their inability to interpret informal multimodal inputs. To bridge this gap, modern frameworks~\citep{wu2024egps,peng2023geodrl} like Inter-GPS~\citep{lu2021intergps} and UniGeo~\citep{cheng2025geouniunifiedmodelgenerating} integrate deep learning parsers with symbolic solvers. However, these systems predominantly rely on a decoupled paradigm where text and diagrams are processed by isolated neural modules. Recent advancements, notably Pi-GPS~\citep{Zhao_2025_ICCV} and AutoGPS~\citep{autogps}, attempt to mitigate this isolation by introducing specialized post-hoc rectification stages or diagrammatic heuristics to resolve referential ambiguities. Despite these efforts, such systems remain tethered to the quality of initial independent parsing results. This structural separation often leads to cascading information loss, as the formalization process lacks a unified representation to resolve inter-modal dependencies, particularly when faced with complex spatial relationships that are under-specified in text or diagram.

\noindent 
\textbf{Multimodal autoformalization.}
Autoformalization bridges the gap between informal human intuition and machine-verifiable logic~\citep{NEURIPS2022_d0c6bc64}. In geometry, this task is particularly formidable as linguistic descriptions are often logically under-specified without visual grounding. The significance of this field is underscored by MATP-BENCH~\citep{he2025matpbenchmllmgoodautomated}, which highlights that formally verified multimodal mathematics remains a grand challenge for frontier MLLMs. Recent frameworks, such as AutoGPS~\citep{autogps} and Pi-GPS~\citep{Zhao_2025_ICCV}, attempt to address this by translating diagrams into predicates. However, they predominantly rely on a decoupled paradigm that utilizes specialized, non-transformer parsers like PGDP~\citep{hao2022pgdp5k}, necessitating post-hoc heuristics to resolve inter-modal inconsistencies. Similarly, DFE-GPS~\citep{zhang2024diagramformalizationenhancedmultimodal} treats diagram formalization merely as an auxiliary signal rather than a primary objective. While recent MMFORMALIZER~\citep{xiong2026mmformalizermultimodalautoformalizationwild} extends autoformalization to physics via recursive grounding and rule-based searches in PhysLean, it focuses on high-level axiomatic deduction. In contrast, our framework achieves joint diagram-text alignment within a unified LMM representation. By enabling bidirectional disambiguation in a single inference pass, we resolve low-level topological ambiguities directly from raw pixels, offering a more integrated and scalable pathway for verifiable multimodal reasoning.


\noindent
\textbf{Verifiable reinforcement learning.}
Reinforcement learning has also been explored in geometry reasoning, for example GeoDRL formulates deductive reasoning as sequential decision making over geometry logic graphs~\citep{peng2023geodrl}. More recently, LLM reasoning has shifted toward \emph{verifiable} rewards~\citep{deepseekai2025deepseekr1incentivizingreasoningcapability}, where correctness can be computed automatically from math or code outputs instead of relying on human preference labels. DeepSeekMath popularized this paradigm in mathematical reasoning with GRPO~\citep{grpo}, while GSPO later improved training stability and efficiency through sequence-level policy optimization~\citep{zheng2025groupsequencepolicyoptimization}. In contrast, our work anchors optimization on solver-readiness of the formal representation. We adapt GSPO to multimodal autoformalization with rewards derived from parse validity, executable solver states, and answer-level solvability. When gold formal annotations are available, we use them for supervised adaptation and analysis; the reinforcement-learning stage itself can be driven by parser and solver execution feedback, making the objective closer to the downstream symbolic task.


\section{Method}
\label{sec:method}

\subsection{Framework Overview}
\label{subsec:overview}

Given a geometry problem consisting of a diagram $I$, a textual question $T$, and a target answer $a^\ast$, \ours aims to generate a formal representation $\hat{\Lambda}$ that can be consumed by a symbolic solver $\mathcal{S}$. The system is organized around two solver-coupled loops. The first loop learns a multimodal formalizer $F_\theta$:
\begin{equation}
    \hat{\Lambda} = F_\theta(I,T),
\end{equation}
where $\hat{\Lambda}$ contains formal predicates grounded in both the diagram and the text. The second loop augments symbolic deduction with a theorem-proposing agent. When $\mathcal{S}(\hat{\Lambda})$ cannot derive the target answer, the agent observes the intermediate proof state and proposes a theorem hypothesis $\hat{\tau}$, after which the solver verifies whether $\hat{\tau}$ is applicable and useful for continuing the proof. This design keeps all accepted formalizations and theorem proposals grounded in symbolic execution.

\begin{figure}[t]
\centering
\includegraphics[width=0.98\textwidth]{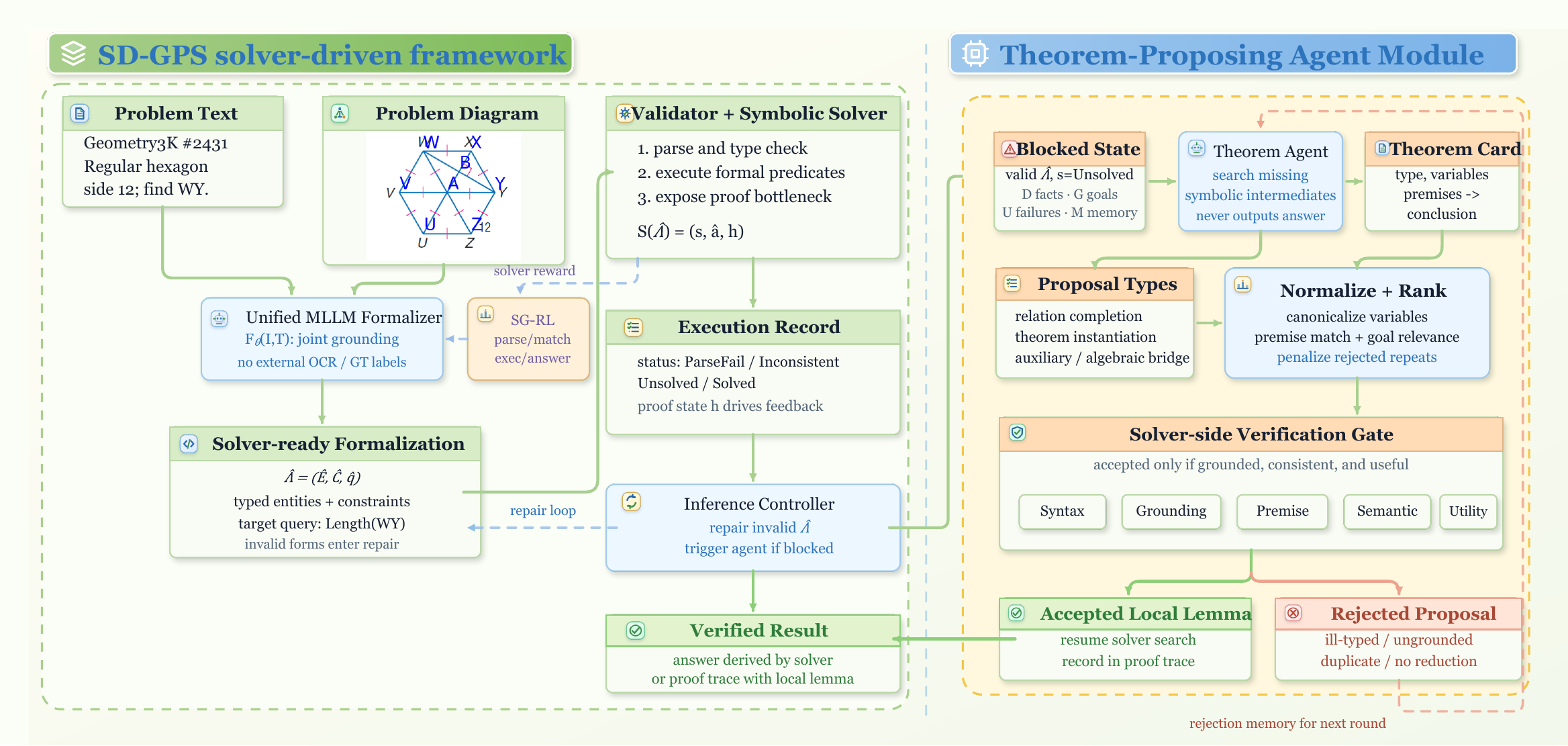}
\caption{Overview of the proposed \ours{} framework. Raw diagram-text inputs are converted into solver-executable formal predicates, checked by a symbolic solver, refined through solver-guided repair when necessary, and augmented with verified auxiliary theorem proposals when the solver reaches a deductive impasse.}
\label{fig:framework_overview}
\end{figure}

Figure~\ref{fig:framework_overview} illustrates the overall workflow of \ours, including multimodal formalization, solver feedback, repair, and theorem proposal.

\subsection{Formalization Data: Curated and Synthetic Sources}
\label{subsec:data_curation}

To expose the formalizer to realistic textbook phrasing, diverse diagram styles, and annotation noise, we build a mixed formalization corpus from curated real-world examples and controllable synthetic instances. The real-world portion is constructed from Geometry3K-Train, the training split of PGPS9K, and the PGDP training set. Since these datasets differ in annotation conventions and were not originally designed as a unified strict formal-language benchmark, we normalize symbolic expressions, entity names, and diagram-text grounding formats. Geometry3K annotations are proofread and standardized to reduce inconsistent predicate usage and ambiguous entity references. For PGPS9K, diagram-side formal language is converted into the Inter-GPS formalism, disambiguated during preprocessing, and then filtered by parser and solver execution. Samples whose converted predicates fail parsing, entity grounding, or solver verification are removed.

The synthetic portion is generated from geometry templates that specify a diagram construction program, a predicate-level formalization, and a target query. Template parameters control point layout, label visibility, auxiliary marks, algebraic values, and ambiguity type. We use the solver to reject inconsistent constructions and keep only examples whose predicates are executable and whose target answer is uniquely determined. This gives the formalizer additional supervision for cases that are underrepresented in real data, such as omitted labels, diagram-supplied points, dense angle marks, and visually similar point names.

The final training data are tuples $(I,T,\Lambda^\ast,a^\ast,m)$, where $I$ denotes the diagram, $T$ denotes the problem text, $\Lambda^\ast$ denotes the target formal representation, $a^\ast$ denotes the gold answer, and $m$ stores metadata about entity grounding, ambiguity type, annotation source, and solver status. This metadata supports curriculum sampling. Early training emphasizes clean, solver-verified examples with direct text-diagram alignment, while later training gradually increases the proportion of diagram-dependent, ambiguity-heavy, and cross-modal-mismatch examples. We keep training, validation, and test problems separated at the problem and diagram levels to reduce leakage between curated or template-derived instances and held-out evaluation examples.

\subsection{Solver-Driven Autoformalization}
\label{subsec:autoformalization}

Solver-Driven Autoformalization is the front-end loop that converts raw multimodal inputs into a formal representation that the solver can execute. Rather than first extracting OCR tokens, diagram primitives, and textual predicates through separate modules, the formalizer $F_\theta$ receives the raw diagram $I$ and problem text $T$ jointly and emits a complete predicate sequence $\hat{\Lambda}$ in one pass. A lightweight parser checks arity, type compatibility, duplicated declarations, illegal symbols, and unresolved references before any candidate is passed to the symbolic solver $\mathcal{S}$. In our implementation, $F_\theta$ is initialized from \backbone~\citep{bai2025qwen3vltechnicalreport}; all supervised, RL, repair, and theorem-proposal variants use this same 2B-parameter multimodal backbone unless explicitly stated otherwise. The same parser and solver are then used in both training and inference, so the model is optimized toward the interface actually consumed by the downstream reasoner.

\paragraph{Stage I: supervised formal-language adaptation.}
Given the curated and synthetic formalization dataset
\begin{equation}
\mathcal{D}_{\mathrm{SFT}}
=
\left\{(I_i, T_i, \Lambda_i^\ast, a_i^\ast)\right\}_{i=1}^{N},
\end{equation}
we first adapt \backbone, a 2B-parameter pretrained multimodal language model, to the target geometry formal language. The model generates
\begin{equation}
\hat{\Lambda}_i = F_\theta(I_i, T_i),
\end{equation}
and is trained with the autoregressive negative log-likelihood of the target predicate sequence $\Lambda_i^\ast=(y_{i,1},\ldots,y_{i,L_i})$:
\begin{equation}
\mathcal{L}_{\mathrm{SFT}}(\theta)
=
-\frac{1}{N}
\sum_{i=1}^{N}
\sum_{t=1}^{L_i}
\log p_\theta
\left(
 y_{i,t}\mid I_i,T_i,y_{i,<t}
\right).
\end{equation}
This stage teaches the predicate inventory, output grammar, and cross-modal grounding patterns, providing a solver-compatible initialization before execution-based optimization.

\paragraph{Stage II: solvability-guided policy optimization.}
After supervised adaptation, we further optimize $F_\theta$ with Solvability-Guided Reinforcement Learning (SG-RL), instantiated with GSPO-style group sequence policy optimization. For each problem, the model samples a group of $K$ candidate formalizations $\{\hat{\Lambda}_{i,j}\}_{j=1}^{K}$. Each candidate is parsed and, if valid, executed by $\mathcal{S}$. We assign a sequence-level reward
\begin{equation}
R(\hat{\Lambda})
=
\lambda_{\mathrm{fmt}}R_{\mathrm{fmt}}(\hat{\Lambda})
+
\lambda_{\mathrm{exec}}R_{\mathrm{exec}}(\hat{\Lambda})
+
\lambda_{\mathrm{ans}}R_{\mathrm{ans}}(\hat{\Lambda},a^\ast),
\end{equation}
where
\begin{equation}
R_{\mathrm{fmt}}(\hat{\Lambda})=\mathbb{I}[\mathrm{Parse}(\hat{\Lambda})=1],
\end{equation}
\begin{equation}
R_{\mathrm{exec}}(\hat{\Lambda})=\mathbb{I}[\mathcal{S}\ \text{constructs a non-contradictory proof state from }\hat{\Lambda}],
\end{equation}
and
\begin{equation}
R_{\mathrm{ans}}(\hat{\Lambda},a^\ast)=\mathbb{I}[\mathcal{S}(\hat{\Lambda})=a^\ast].
\end{equation}
The format term rewards syntactic validity, the execution term rewards usable intermediate solver states, and the answer term rewards candidates that are sufficient for deriving the gold answer. Invalid or unparsable candidates receive no execution or answer reward. Within each sampled group, rewards are normalized into sequence-level advantages and used to update the policy, so formalizations that are merely plausible but unusable by the solver are discouraged. Unlike the supervised stage, SG-RL does not require gold formal-language annotations for every problem; it requires the problem input, the target answer, and solver feedback.

\paragraph{Inference-time use.}
At test time, the same parser-solver interface provides structured diagnostics. If a candidate fails parsing or execution, it is sent to the bounded repair loop described in Appendix~\ref{app:verification_repair}. If the candidate is executable but the solver cannot close the proof within the search budget, the system invokes the theorem-proposing agent in Section~\ref{subsec:agent}.

\subsection{Solver-Driven Theorem-Proposing Agent}
\label{subsec:agent}

Even a correct and executable formalization may be insufficient when the symbolic solver is restricted to a fixed theorem library or a finite search budget. The Solver-Driven Theorem-Proposing Agent addresses this bottleneck by acting as a proposal module over the current proof state, not as an unchecked source of new axioms. In our implementation, the agent is driven by the same QwenVL3-2B backbone as the formalizer, but its output is restricted to a verifier-readable proposal schema. It is activated only after $\mathcal{S}$ has successfully parsed $\hat{\Lambda}$ and built a consistent proof state but cannot derive the target query $q$. At iteration $t$, the solver exposes a state summary $h_t$ containing known facts, unresolved subgoals, failed theorem matches, and a compact diagnostic message $d_t$. Conditioned on $(\hat{\Lambda},q,h_t,d_t)$, the agent proposes a bounded set of candidate auxiliary lemmas
\begin{equation}
\mathcal{T}_t = A_\phi(\hat{\Lambda}, q, h_t, d_t),
\end{equation}
where each candidate is represented as a verifier-readable tuple containing a lemma type, required premises, proposed conclusion, involved entities, and a short rationale.

The solver filters these candidates before any of them can affect the derivation. A proposal is discarded if it contains unknown entities, violates type constraints, contradicts existing facts, cannot be instantiated against the current proof state, cannot be verified from admissible premises, or fails to reduce any unresolved subgoal. For an accepted proposal, the verifier returns a local proof certificate or a deterministic construction trace, and the derived fact is scoped only to the current problem instance. It is removed after the example is solved or declared unresolved. This scoping is important: the agent may suggest a useful lemma instantiation or auxiliary relation, but the symbolic verifier remains responsible for deciding whether the suggestion is admissible. In practice, the agent mostly helps in three cases: selecting a theorem instance that the default search failed to prioritize, exposing a latent relation such as cyclicity or similarity, and proposing a small auxiliary construction that makes an existing theorem applicable.

\begin{algorithm}[t]
\caption{Solver-driven theorem proposal and verification}
\label{alg:theorem_agent}
\begin{algorithmic}[1]
\Require Formal predicates $\hat{\Lambda}$, target query $q$, solver $\mathcal{S}$, theorem agent $A_\phi$, maximum proposal rounds $T_{\max}$, proposal budget $K$
\Ensure Solved answer or unresolved status
\State $(s_0,h_0) \gets \mathcal{S}.\textsc{Initialize}(\hat{\Lambda},q)$
\If{$s_0$ is invalid or contradictory}
    \State \Return unresolved \Comment{Handled by formalization repair, not theorem proposal}
\EndIf
\If{$\mathcal{S}.\textsc{Closed}(s_0,q)$}
    \State \Return $\mathcal{S}.\textsc{Answer}(s_0,q)$
\EndIf
\For{$t=1$ to $T_{\max}$}
    \State $s_t \gets s_{t-1}$
    \State $d_t \gets \mathcal{S}.\textsc{Diagnose}(s_{t-1},q)$
    \State $\mathcal{T}_t \gets A_\phi(\hat{\Lambda},q,h_{t-1},d_t;K)$
    \ForAll{$\tau \in \mathcal{T}_t$}
        \If{not $\textsc{WellTyped}(\tau,\hat{\Lambda})$}
            \State discard $\tau$
        \ElsIf{not $\mathcal{S}.\textsc{Applicable}(\tau,s_{t-1})$}
            \State discard $\tau$
        \ElsIf{not $\mathcal{S}.\textsc{Verify}(\tau,s_{t-1})$}
            \State discard $\tau$
        \ElsIf{not $\mathcal{S}.\textsc{GoalReduced}(\tau,s_{t-1},q)$}
            \State discard $\tau$
        \Else
            \State $(s_t,c_\tau) \gets \mathcal{S}.\textsc{AddLocalLemma}(s_{t-1},\tau)$
            \State $s_t \gets \mathcal{S}.\textsc{Search}(s_t,q)$
            \If{$\mathcal{S}.\textsc{Closed}(s_t,q)$}
                \State \Return $\mathcal{S}.\textsc{Answer}(s_t,q)$
            \EndIf
        \EndIf
    \EndFor
    \State $h_t \gets \mathcal{S}.\textsc{Summarize}(s_t,q)$
\EndFor
\State \Return unresolved
\end{algorithmic}
\end{algorithm}

Algorithm~\ref{alg:theorem_agent} summarizes the loop. The proposal budget keeps inference bounded, while the verifier checks ensure that neural proposals cannot directly determine the answer. The stored certificate $c_\tau$ is used only inside the current proof attempt, so accepted proposals behave as verified local lemmas rather than permanent changes to the theorem library. The agent therefore expands the solver's search behavior without replacing symbolic proof checking.

\section{Experiments}
\label{sec:expr}
\begin{table}[t]
\centering
\small
\setlength{\tabcolsep}{6pt}
\renewcommand{\arraystretch}{1.2}
\caption{Results on Geometry3K and PGPS9K.}
\label{tab:geo3k_pgps9k_grouped}
\begin{tabular}{l cc cc}
\toprule
\textbf{Method} 
& \multicolumn{2}{c}{\textbf{Geometry3K}} 
& \multicolumn{2}{c}{\textbf{PGPS9K}} \\
\cmidrule(lr){2-3}\cmidrule(lr){4-5}
& \textbf{Completion} & \textbf{Choice} & \textbf{Completion} & \textbf{Choice} \\
\midrule
\multicolumn{5}{c}{\textbf{\textit{MLLMs}}} \\
G-LLaVA-13B~\citep{gao2025gllava} & 0.3 & 29.0 & 0.0 & 27.0 \\
Vision-R1-7B~\citep{huang2025visionr1} & 43.8 & 57.1 & 36.8 & 49.6 \\
Qwen2.5-VL-32B~\citep{qwen2025qwen25technicalreport} & 41.9 & 67.6 & 43.5 & 56.1\\
InternVL2.5-78B~\citep{chen2024internvl} & 36.1 & 60.9 & 28.1 & 51.3 \\
InternVL3-78B~\citep{zhu2025internvl3} & 57.4 & 74.5 & 48.9 & 61.1\\
GPT-4o~\citep{openai2024gpt4ocard} & 34.8 & 58.6 & 38.2 & 56.8 \\
GLM-4.6V~\citep{vteam2026glm45vglm41vthinkingversatilemultimodal} & 53.7 & 65.3 & 47.7 & 60.8 \\
Kimi-K2.5~\citep{kimiteam2026kimik25visualagentic} & 82.9 & 87.2 & 73.0 & 79.8 \\
QwenVL3-2B~\citep{bai2025qwen3vltechnicalreport} & 36.6 & 52.5 & 27.2 & 45.4 \\
\midrule
\multicolumn{5}{c}{\textbf{\textit{Neural Solvers}}} \\
NGS~\citep{chen2022geoqageometricquestionanswering} & 35.3 & 58.8 & 34.1 & 46.1 \\
Geoformer~\citep{khomiakov2024geoformermultipolygonsegmentationtransformer} & 36.8 & 59.3 & 35.6 & 47.3 \\
SCA-GPS~\citep{ning2023symboliccharacterawaremodelsolving} & 76.7 & -- & -- & -- \\
GOLD~\citep{zhang2024goldgeometryproblemsolver} & -- & 62.7 & -- & 60.6 \\
PGPSNet-v2-S~\citep{zhang2024fusereasonverifygeometry} & 65.2 & 76.4 & 60.3 & 69.2 \\
LANS (Diagram GT)~\citep{li-etal-2024-lans} & 72.1 & 82.3 & 66.7 & 74.0 \\
PGPSNet~\citep{zhang2023pgpsnet} & 65.0 & 77.9 & 62.7 & 70.4\\
\midrule
\multicolumn{5}{c}{\textbf{\textit{Symbolic and Neuro-symbolic Solvers}}} \\
Inter-GPS~\citep{lu2021intergps} & 43.4 & 57.5 & -- & -- \\
E-GPS~\citep{wu2024egps} & -- & 67.9 & -- & -- \\
GeoDRL~\citep{peng2023geodrl} & 57.9 & 68.4 & 55.6 & 66.7 \\
Auto-GPS~\citep{autogps} & 75.4 & 81.6 & 75.4 & 81.5 \\
Auto-GPS (OCR)~\citep{autogps} & 65.6 & 75.2 & 62.9 & 71.9 \\
Pi-GPS~\citep{Zhao_2025_ICCV} & 70.6 & 77.8 & 61.4 & 69.8 \\
Pi-GPS (OCR)~\citep{Zhao_2025_ICCV} & 66.1 & 73.1 & 56.3 & 64.2 \\
\midrule
\ours & \textbf{86.4} & \textbf{90.4} & \textbf{79.8} & \textbf{84.5} \\
\bottomrule
\end{tabular}
\end{table}

\paragraph{Datasets.}
We evaluate on Geometry3K~\citep{lu2021intergps} and PGPS9K~\citep{zhang2023pgpsnet}, the two benchmarks used to test whether solver-driven formalization transfers from raw multimodal input to final problem solving. Geometry3K contains 3,002 geometry problems, split into 2,101 training, 300 validation, and 601 test problems. Each problem includes a diagram, problem text, and formal-language annotations. PGPS9K extends Geometry3K to 9,022 problems paired with 4,000 unique diagrams and covers common plane-geometry types from middle-school and high-school textbooks.

\paragraph{Metrics.}
Following prior GPS evaluations~\citep{lu2021intergps,zhang2023pgpsnet,autogps,Zhao_2025_ICCV}, we report final problem-solving accuracy under two settings. \emph{Completion} evaluates open-ended answer generation, where the system must derive the final value or expression. \emph{Choice} evaluates multiple-choice solving. For ablations on ambiguity resolution, we additionally report final solver-executed accuracy under three text-diagram evidence regimes: \emph{text-complete}, where the text-mentioned point set equals the diagram point set; \emph{diagram-augmented}, where the diagram contains additional points required for grounding; and \emph{cross-modal mismatch}, where the model must resolve weaker or inconsistent entity evidence.

\paragraph{Baselines.}
We compare with three groups of methods that correspond to the failure modes discussed in the introduction. The first group contains general MLLMs prompted to solve or formalize the problem directly, including G-LLaVA~\citep{gao2025gllava}, Vision-R1~\citep{huang2025visionr1}, Qwen2.5-VL~\citep{qwen2025qwen25technicalreport}, QwenVL3-2B~\citep{bai2025qwen3vltechnicalreport}, InternVL~\citep{chen2024internvl,zhu2025internvl3}, and GPT-4o~\citep{openai2024gpt4ocard}. The second group contains neural GPS systems such as NGS~\citep{chen2022geoqageometricquestionanswering}, Geoformer~\citep{khomiakov2024geoformermultipolygonsegmentationtransformer}, SCA-GPS~\citep{ning2023symboliccharacterawaremodelsolving}, GOLD~\citep{zhang2024goldgeometryproblemsolver}, PGPSNet~\citep{zhang2023pgpsnet}, PGPSNet-v2~\citep{zhang2024fusereasonverifygeometry}, and LANS~\citep{li-etal-2024-lans}. The third group contains symbolic or neuro-symbolic solvers such as Inter-GPS~\citep{lu2021intergps}, E-GPS~\citep{wu2024egps}, GeoDRL~\citep{peng2023geodrl}, Auto-GPS~\citep{autogps}, and Pi-GPS~\citep{Zhao_2025_ICCV}. Where available, we separately report OCR-based variants~\citep{autogps,Zhao_2025_ICCV}, because they expose the brittleness of cascaded perception-to-logic pipelines under fully automatic input processing.

\subsection{Results}

Table~\ref{tab:geo3k_pgps9k_grouped} summarizes the results on Geometry3K and PGPS9K. \ours achieves the best accuracy across all four evaluation settings. On Geometry3K, it obtains 86.4\% completion accuracy and 90.4\% choice accuracy, improving over the strongest listed prior results by 3.5 and 3.2 percentage points, respectively. On PGPS9K, \ours reaches 79.8\% completion accuracy and 84.5\% choice accuracy, surpassing the strongest listed prior results by 4.4 and 3.0 percentage points. These gains indicate that solver-driven formalization improves not only predicate quality but also final reasoning performance.

The comparison also highlights the limitations of direct MLLM prompting. Although recent MLLMs perform well on general multimodal tasks, their accuracy remains limited on formal geometry reasoning, especially in completion settings where exact executable reasoning is required. The prompted \backbone baseline is therefore reported separately from \ours: \ours uses the same backbone but adds solver-driven formal-language adaptation, SG-RL, bounded repair, and verified theorem proposal. In contrast to direct prompting, \ours explicitly converts multimodal inputs into solver-ready predicates and uses execution feedback to penalize incomplete or unusable formalizations. The OCR-based variants of Auto-GPS and Pi-GPS further show that cascaded systems are sensitive to upstream recognition errors, reinforcing the need for integrated cross-modal formalization.

\subsection{Ablation Study}

\paragraph{Robustness under cross-modal ambiguity.}
Table~\ref{tab:point_relation_ablation} reports final solver-executed accuracy under different relations between text-mentioned points and diagram points. The SFT-only formalizer performs strongly when the textual point set is complete, but its accuracy decreases in the diagram-augmented and cross-modal-mismatch regimes. This suggests that cross-modal ambiguity is not merely a post-processing issue: missing, redundant, or inconsistently grounded entities can directly change the executable predicates provided to the symbolic solver.

\paragraph{Effect of SG-RL and fallback repair.}
Introducing SG-RL improves the overall accuracy from 74.9\% to 79.5\%, indicating that solver-derived feedback provides a useful supervisory signal beyond predicate-level imitation. The improvement is particularly pronounced in the cross-modal-mismatch setting, where accuracy increases from 67.2\% to 74.8\%. This result suggests that execution feedback helps the model prefer formalizations that are not only syntactically valid but also more compatible with downstream symbolic reasoning. Adding fallback repair further raises the overall accuracy to 84.5\%, with a notable gain in the diagram-augmented setting from 77.1\% to 84.3\%. These results show that bounded execution-aware regeneration can recover from invalid, incomplete, or weakly grounded formalizations.

\paragraph{Effect of theorem proposal.}
The final row of Table~\ref{tab:point_relation_ablation} incorporates the solver-driven theorem-proposing agent, increasing the overall accuracy to 86.4\%. After formalization errors are reduced, some remaining failures are caused by the limited coverage of the fixed theorem library rather than by incorrect parsing alone. By observing the solver state and proposing verifiable auxiliary lemmas, the theorem-proposing agent extends the effective reasoning coverage of the symbolic back-end while preserving verification constraints. Importantly, the proposed lemmas are not accepted as unconstrained neural guesses; they must be checked by the solver before contributing to the final derivation.

\paragraph{Effect of task-specific formalization training.}
Table~\ref{tab:accuracy_without_pre_formalization} compares prompt-based formal-language generation by general-purpose MLLMs with formalizers adapted to the target geometry language. The results show that task-specific adaptation produces more reliable executable formalizations, especially under the syntax, grounding, and predicate constraints imposed by the downstream solver. This supports the need for dedicated formalization training rather than relying solely on prompt-based generation from general-purpose multimodal models.

\begin{table}[t]
\centering
\small
\setlength{\tabcolsep}{4pt}
\renewcommand{\arraystretch}{1.15}
\caption{Fine-grained end-to-end accuracy under different text-diagram point relations. \textit{Text-complete} means the point set mentioned in the problem text equals the diagram point set. \textit{Diagram-augmented} means the text-mentioned point set is a strict subset of the diagram point set. \textit{Cross-modal mismatch} covers the remaining cases.}
\label{tab:point_relation_ablation}
\resizebox{\textwidth}{!}{%
\begin{tabular}{lcccc}
\toprule
\textbf{Variant} & \textbf{Text-complete} ($n=142$) & \textbf{Diagram-augmented} ($n=262$) & \textbf{Cross-modal mismatch} ($n=197$) & \textbf{Overall} ($n=601$) \\
\midrule
Prompted MLLM baseline  & 0.7 & 0.3 & 0 & 0.3 \\
Solver-driven autoformalizer: SFT only & 87.2 & 73.7 & 67.2 & 74.9 \\
\hspace{2em}+ SG-RL solver feedback & 90.8 & 77.1 & 74.8 & 79.5 \\
\hspace{3em}+ bounded fallback repair & 94.3 & 84.3  & 77.7 & 84.5 \\
\hspace{4em}+ theorem-proposing agent  & - & -  & - & 86.4 \\
\bottomrule
\end{tabular}%
}
\end{table}

\begin{table}[htbp]
\centering
\caption{End-to-end solver execution accuracy using model-generated formal representations. All models generate formal representations that are executed by the same symbolic solver. General-purpose MLLMs are evaluated with the same prompt. The \ours{} rows use QwenVL3-2B as the backbone; the QwenVL3-2B row denotes the prompted base model without solver-driven adaptation.}
\begin{tabular}{lc}
\toprule
\textbf{Model Name} & \textbf{Accuracy (\%)} \\ 
\midrule
Prompted QwenVL3-2B    & 0.3   \\
Prompted Qwen2.5-VL-32B   & 18.1  \\
Prompted InternVL3-78B  & 19.3  \\
Prompted GPT-4o       & 22.6  \\
Prompted GLM-4.6V       & 22.6  \\
Prompted Kimi-K2.5      & 28.5 \\ 
\midrule
\textbf{Ours (QwenVL3-2B + SFT)}           & \textbf{74.9} \\
\textbf{Ours (QwenVL3-2B + SG-RL)}          & \textbf{79.5} \\ 
\bottomrule
\end{tabular}

\label{tab:accuracy_without_pre_formalization}
\end{table}

\section{Limitations}
\label{sec:limitations}
Our evaluation is limited to plane-geometry problems that can be represented with an Inter-GPS-style formal language. Extending the framework to three-dimensional geometry, construction-based problems, or free-form proof generation may require a richer predicate vocabulary and additional symbolic verification rules.

The method also depends on the coverage of the symbolic solver. Solver feedback is useful for training and verification, but solver failure may arise from missing theorem rules or limited search depth rather than incorrect formalization. Similarly, the theorem-proposing agent is constrained by the verifier: its proposed auxiliary relations are accepted only when they pass symbolic checks and help reduce the unresolved goal. The agent should therefore be interpreted as a bounded proposal mechanism for theorem instances and auxiliary relations, not as a replacement for formal proof certification.
\section{Conclusion}

We presented \ours, a solver-driven framework for verifiable geometry problem solving. Instead of treating autoformalization as a static translation task and theorem prediction as a closed-library selection problem, \ours uses symbolic execution as an active supervisory signal. A unified multimodal formalizer produces solver-ready predicates from raw diagram-text inputs, solver-verified reinforcement learning aligns training with downstream executability, and an iterative theorem-proposing agent expands the solver's deductive reach through verified lemma proposals. Experiments on Geometry3K and PGPS9K show consistent improvements over MLLM, neural, and neuro-symbolic baselines. These results suggest that closing the loop between perception and symbolic verification is a promising direction for robust multimodal mathematical reasoning. Future work will extend the framework to broader formal languages, stronger proof-certificate generation, and more diverse geometric domains beyond plane-geometry benchmarks.

\bibliographystyle{unsrtnat}
\bibliography{nipis_conference}


\appendix

\section{Technical appendices and supplementary material}
\subsection{Experimental and implementation details}
The symbolic method Inter-GPS~\citep{lu2021intergps} and Auto-GPS~\citep{autogps} were reproduced using the authors' open-source code, while results for
GeoDRL~\citep{peng2023geodrl} and E-GPS~\citep{wu2024egps} were extracted from original publications
due to code unavailability. All experiments were executed on an Intel Xeon Gold 6530 CPU
 platform in conjunction with two NVIDIA L40S GPUs. The trainable multimodal
formalizer is initialized from \backbone; the prompted
\backbone rows use the same base model without task-specific adaptation. A strict timeout
threshold of 600 seconds was imposed on symbolic solvers, where any computation exceeding this
duration was systematically categorized as resolution failure. For MLLM hyperparameters, we set
\texttt{max\_tokens}=4096, \texttt{temperature}=0.1, and \texttt{top-p}=1.0.

\paragraph{Backbone.}
The \ours{} formalizer is initialized from QwenVL3-2B, and all reported \ours{} variants use this 2B multimodal backbone unless otherwise specified. The theorem-proposing agent uses Qwen3.6-Plus. The QwenVL3-2B entries in the baseline tables denote the prompted base model without solver-driven adaptation.

\paragraph{Input normalization.}
For each example, the diagram is resized while preserving aspect ratio and is paired with the raw
problem statement. We do not expose OCR tokens, detected bounding boxes, or manually curated point
sets to \ours during inference. Text is normalized only for Unicode variants of mathematical
symbols and whitespace; numerical values, point labels, and algebraic expressions are otherwise
kept unchanged to avoid leaking solver-side corrections into the formalizer.

\paragraph{Formal representation.}
We adopt an InterGPS-style typed formal language to represent geometric problems in a solver-executable form.
The formalizer emits a structured sequence of typed predicates, explicitly separating entity declarations, relational constraints, and query targets. Each problem instance is represented as a directed set of well-typed logical atoms over geometric primitives.
This InterGPS-style representation enforces strict type consistency between geometric objects and operators, preventing invalid compositions such as applying length functions to angles or mixing segment-level and point-level relations.
Before being passed to the solver, a lightweight structural parser performs validation of:
arity constraints, type compatibility, duplicate entity declarations, illegal symbol usage, and unresolved references.

\subsection{Inference-time verification and repair}
\label{app:verification_repair}

At inference time, \ours uses a bounded verify-repair loop. The first pass generates a candidate
formalization. The parser and solver then return structured diagnostics, including missing entity
references, ill-typed predicates, contradictory constraints, unsolved targets, and timeout states.
If the candidate fails before a valid proof state is constructed, the model is prompted to repair
only the formal predicates while preserving the original problem image and text. We allow at most
two repair attempts per example. The final answer is accepted only when it is produced from a
solver-executable formalization; otherwise the example is marked unresolved. This policy keeps the
reported accuracy tied to verifiable execution rather than unconstrained natural-language
rationales.

\subsection{Three-way split of point-reference ambiguity}

To analyze how much information is explicitly available from the problem text, we divide examples
according to the relation between text-mentioned point names and diagram point instances. We first
extract the point set $P_{\mathrm{text}}$ that is explicitly mentioned in the problem statement.
The extractor is constrained by the diagram point vocabulary $P_{\mathrm{diag}}$, so it can only
return valid diagram point names and cannot infer unmentioned points from geometry or common sense.
The split is then defined as follows:
\begin{itemize}
    \item \textbf{Text-complete:} $P_{\mathrm{text}} = P_{\mathrm{diag}}$. The text explicitly
    names all diagram points needed by the visual instance.
    \item \textbf{Diagram-augmented:} $P_{\mathrm{text}} \subset P_{\mathrm{diag}}$. The text names
    only a strict subset of diagram points, so the diagram supplies additional entities required for
    formalization.
    \item \textbf{Cross-modal mismatch:} all remaining cases, including examples where the text
    refers to an underspecified figure, region, or target whose executable meaning must be resolved
    from the diagram and the formal constraints together.
\end{itemize}

\begin{figure*}[t]
\centering
\includegraphics[width=0.98\textwidth]{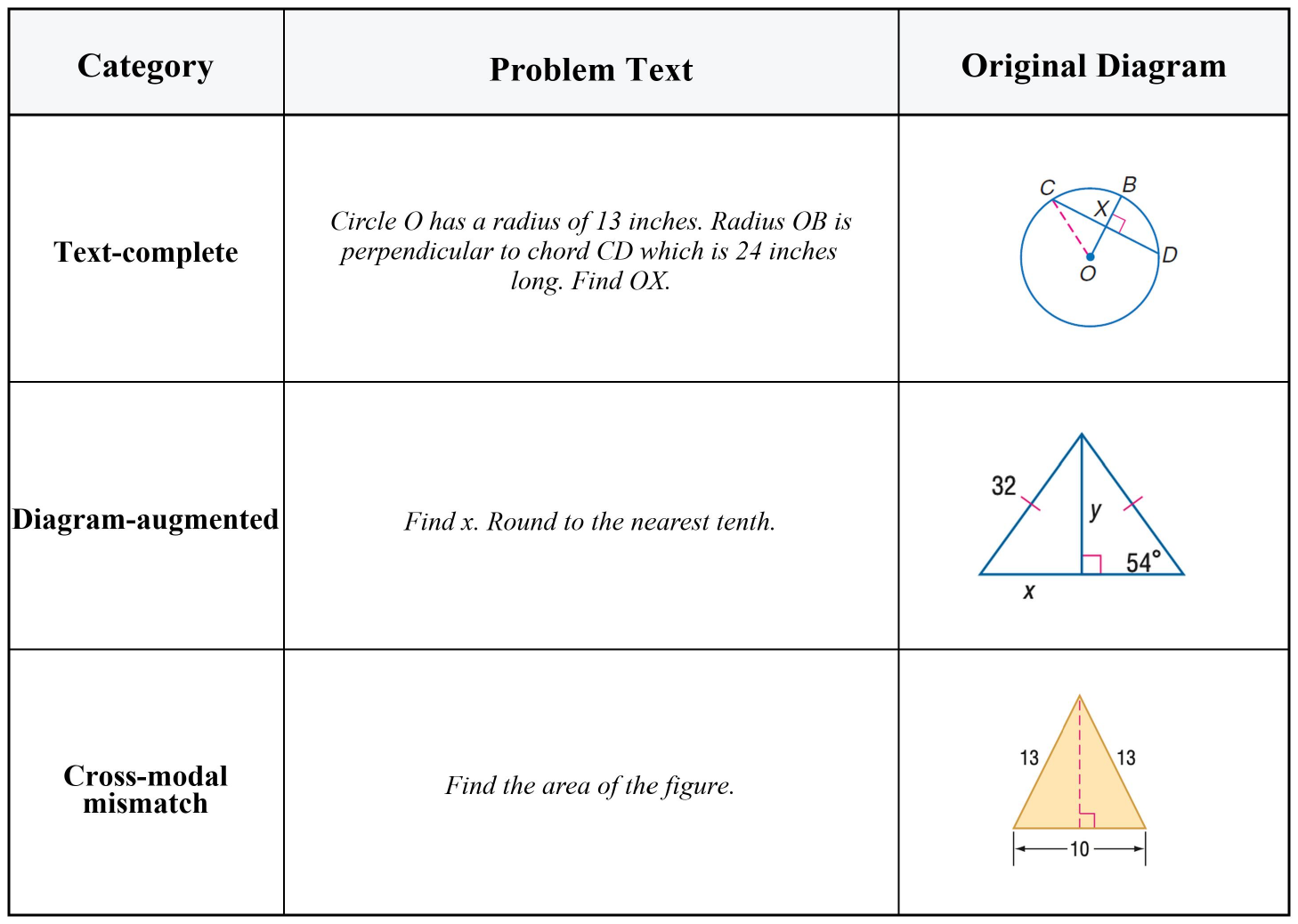}
\caption{Representative examples for the three-way split of point-reference ambiguity. The
classification compares point names explicitly recoverable from the problem text with point
instances present in the diagram, then evaluates whether the resulting formal language can resolve
the target query.}
\label{fig:category_split_examples}
\end{figure*}

Figure~\ref{fig:category_split_examples} shows one representative example from each split. These
examples are used only for qualitative illustration; the quantitative ablation in
Table~\ref{tab:point_relation_ablation} is computed over the full test split.

\subsection{Additional theorem-proposing agent details}

The theorem-proposing agent is invoked only after the formalization is executable but the solver
cannot close the proof within the timeout or theorem-search budget. The agent observes four
signals: the normalized predicate set, the current solver agenda, the unresolved target, and the
latest failure diagnostic. We serialize this information as a compact proof-state summary rather
than a full search tree, which keeps the prompt short and prevents the model from relying on
irrelevant derivations.

Each proposal follows a restricted schema:
\begin{equation}
\tau=(\mathrm{type},\mathrm{premises},\mathrm{conclusion},\mathrm{entities},\mathrm{rationale}).
\end{equation}
The type field specifies whether the proposal is, for example, a similarity relation, cyclicity
relation, equal-angle relation, proportionality relation, or auxiliary construction. The entity
field must use only point, line, circle, or expression identifiers already declared in
$\hat{\Lambda}$, unless the proposal type explicitly permits a bounded auxiliary construction.
This schema makes the proposal easy to parse and prevents free-form natural-language suggestions
from entering the solver.

A proposal is accepted only if it passes four checks: well-typedness under the formal language,
applicability to the current proof state, symbolic verification of the proposed conclusion from
admissible premises, and goal reduction under the solver's search procedure. Unverified proposals
are discarded and do not affect the final answer. Verified proposals are treated as temporary
lemmas scoped to the current problem instance rather than permanent additions to the global theorem
library, and the verifier stores the corresponding local certificate or construction trace. This
design allows the system to bridge instance-specific deductive gaps while preserving the solver's role as the authority for correctness.

\subsection{OCR failure modes in cascaded pipelines}

The gap between fully automatic OCR-based pipelines and their non-OCR counterparts is not only a matter of small transcription noise. In geometry diagrams, labels are often tiny, slanted, partially occluded by arrows or tick marks, and visually entangled with nearby symbolic marks. Under these conditions, OCR errors frequently alter the symbolic identity of an entity or the algebraic form of a constraint, which then propagates directly into the formal language used by the downstream solver.

\begin{figure*}[t]
\centering
\includegraphics[width=0.98\textwidth]{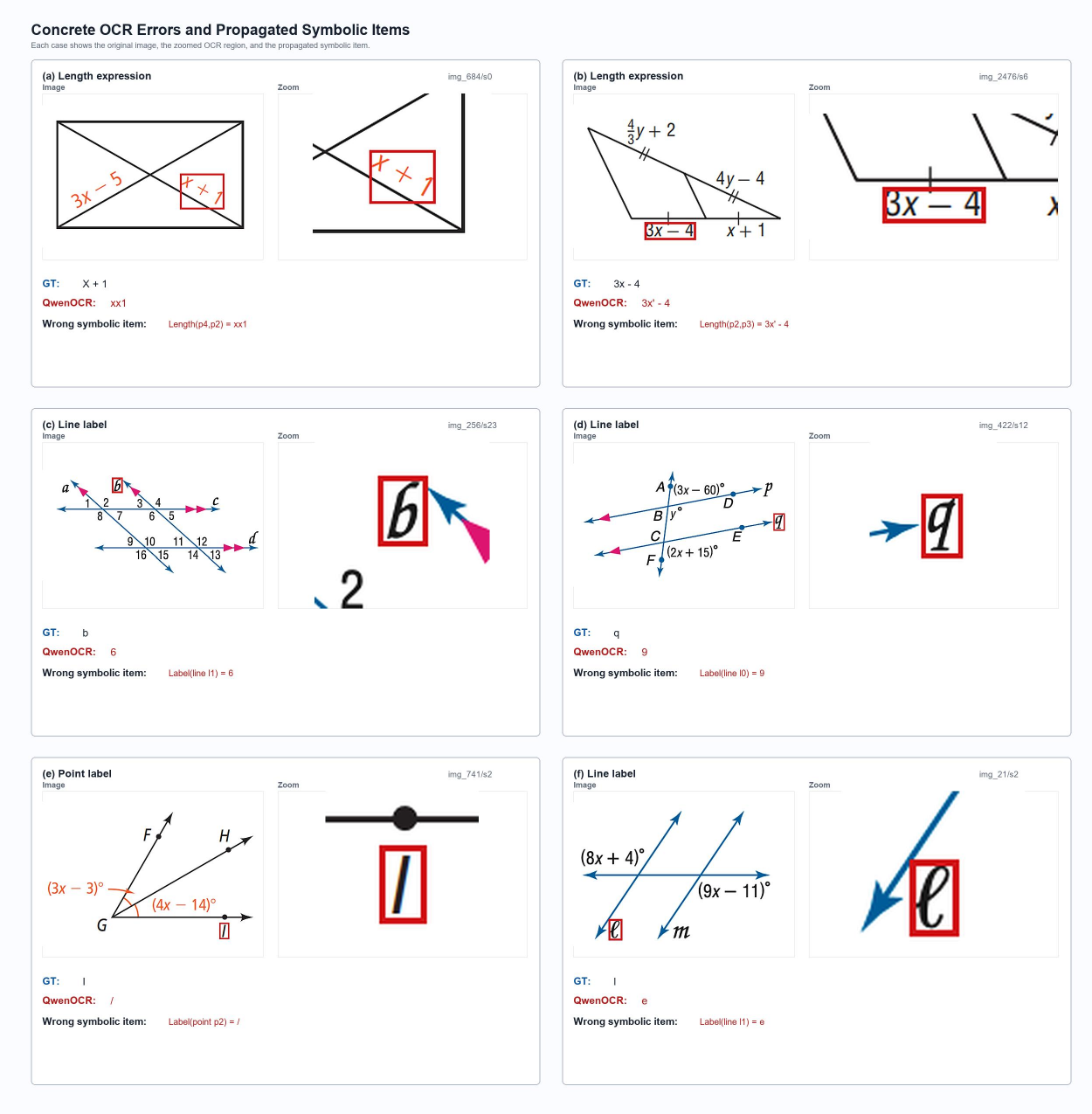}
\caption{Representative OCR failures and their propagated symbolic consequences in cascaded geometry pipelines. Panels (a)--(b) show expression-level corruption, including operator deletion and spurious stroke insertion. Panels (c)--(f) show label-level confusion, where small line or point labels are misread as digits or visually similar glyphs. In each case, the OCR output is not merely cosmetically wrong; it induces an incorrect symbolic item that changes the formal problem passed to the downstream solver.}
\label{fig:ocr_error_cases}
\end{figure*}

Figure~\ref{fig:ocr_error_cases} shows representative OCR failures on geometry diagram crops. The first two cases illustrate expression-level corruption: in panel (a), the length expression \texttt{x + 1} is read as \texttt{x1}, effectively deleting the operator; in panel (b), \texttt{3x - 4} is misread as \texttt{3x' - 4}, injecting an extra stroke into the symbolic item. The remaining four cases illustrate label-level confusion. Lowercase letters can be mapped to digits or visually similar glyphs, e.g., \texttt{b} $\rightarrow$ \texttt{6} in panel (c), \texttt{q} $\rightarrow$ \texttt{9} in panel (d), the point label \texttt{I} $\rightarrow$ \texttt{/} in panel (e), and the line label \texttt{l} $\rightarrow$ \texttt{e} in panel (f).

These failures are especially damaging in cascaded geometry systems because OCR output is typically treated as an authoritative symbol table for subsequent parsing and reasoning. Once a segment length is written as \texttt{x1} instead of \texttt{x + 1}, or a line label is changed from \texttt{l} to \texttt{e}, downstream modules no longer operate on a slightly noisy observation; they operate on a different formal problem. The symbolic solver can verify consistency only with respect to the corrupted predicates it receives, so it has very limited ability to recover from these front-end mistakes. This explains why OCR-based variants in the main results degrade much more sharply than end-to-end approaches that avoid explicit OCR as an intermediate bottleneck.

\end{document}